\documentclass{article}

\usepackage{microtype}
\usepackage{graphicx}
\usepackage{subcaption}
\usepackage{booktabs} % for professional tables

\usepackage{hyperref}

\usepackage[preprint]{icml2026}

\usepackage{amsmath}
\usepackage{amssymb}
\usepackage{mathtools}
\usepackage{amsthm}

\usepackage{adjustbox}
\usepackage{tabularx}
\usepackage{caption}
\usepackage{multirow}
\usepackage{pifont}
\usepackage{enumitem}
\usepackage{makecell}
\usepackage{xcolor}

\newcommand{\red}[1]{\textcolor{red}{#1}}
\newcommand{\green}[1]{\textcolor{green!60!black}{#1}}

\newcommand{\promptfield}[1]{\textcolor{blue}{$<$#1$>$}}
\definecolor{softgreen}{rgb}{0.0, 0.5, 0.0}
\definecolor{softred}{rgb}{0.7, 0.0, 0.0}

\usepackage[capitalize,noabbrev]{cleveref}

\theoremstyle{plain}

\theoremstyle{definition}

\theoremstyle{remark}

\usepackage[textsize=tiny]{todonotes}

\makeatletter
\renewcommand{\printAffiliationsAndNotice}[1]{\global\icml@noticeprintedtrue%
  \stepcounter{@affiliationcounter}%
  {\let\thefootnote\relax\footnotetext{\hspace*{-\footnotesep}\ificmlshowauthors #1\fi%
      \forloop{@affilnum}{1}{\value{@affilnum} < \value{@affiliationcounter}}{
        \textsuperscript{\arabic{@affilnum}}\ifcsname @affilname\the@affilnum\endcsname%
        \csname @affilname\the@affilnum\endcsname%
        \else
        {\bf AUTHORERR: Missing \textbackslash{}icmlaffiliation.}
        \fi
      }.%

      \ \\
      \Notice@String
    }
  }
}
\makeatother

\icmltitlerunning{Revisiting Graph-Tokenizing Large Language Models: A Systematic Evaluation of Graph Token Understanding}
\begin{document}
\raggedbottom

\twocolumn[
  % \icmltitle{Revisiting Graph-Tokenizing Large Language Models: How Do They Understand Graph Tokens?}
  \icmltitle{Revisiting Graph-Tokenizing Large Language Models: A Systematic Evaluation of Graph Token Understanding}

  \icmlsetsymbol{equal}{*}
  \icmlsetsymbol{cor}{\textdagger}

  % \begin{icmlauthorlist}
  %   \icmlauthor{Firstname1 Lastname1}{equal,yyy}
  %   \icmlauthor{Firstname2 Lastname2}{equal,yyy,comp}
  %   \icmlauthor{Firstname3 Lastname3}{comp}
  %   \icmlauthor{Firstname4 Lastname4}{sch}
  %   \icmlauthor{Firstname5 Lastname5}{yyy}
  %   \icmlauthor{Firstname6 Lastname6}{sch,yyy,comp}
  %   \icmlauthor{Firstname7 Lastname7}{comp}
  %   %\icmlauthor{}{sch}
  %   \icmlauthor{Firstname8 Lastname8}{sch}
  %   \icmlauthor{Firstname8 Lastname8}{yyy,comp}
  %   %\icmlauthor{}{sch}
  %   %\icmlauthor{}{sch}
  % \end{icmlauthorlist}

  \begin{icmlauthorlist}
    \icmlauthor{Zhongjian Zhang}{equal,bupt}
    \icmlauthor{Yue Yu}{equal,bupt}
    \icmlauthor{Mengmei Zhang}{bestpay}
    \icmlauthor{Junping Du}{bupt}
    \icmlauthor{Xiao Wang}{cor,buaa}
    \icmlauthor{Chuan Shi}{cor,bupt}
  \end{icmlauthorlist}

  \icmlaffiliation{bupt}{Beijing University of Posts and Telecommunications}
  \icmlaffiliation{bestpay}{China Telecom Bestpay}
  \icmlaffiliation{buaa}{Beihang University}

  \icmlcorrespondingauthor{Xiao Wang}{xiao\_wang@buaa.edu.cn}
  \icmlcorrespondingauthor{Chuan Shi}{shichuan@bupt.edu.cn}

  \icmlkeywords{Machine Learning, ICML}

  \vskip 0.3in
]

% this must go after the closing bracket ] foPreliminary work. Under review by the International llowing \twocolumn[ ...


\printAffiliationsAndNotice{\icmlEqualContribution\textsuperscript{\textdagger}Corresponding authors }  % warning from hyperref is expected

\begin{abstract}
  The remarkable success of large language models (LLMs) has motivated researchers to adapt them as universal predictors for various graph tasks.
  As a widely recognized paradigm, Graph-Tokenizing LLMs (GTokenLLMs) compress complex graph data into graph tokens and treat them as prefix tokens for querying LLMs, leading many to believe that LLMs can understand graphs more effectively and efficiently. In this paper, we challenge this belief: \textit{Do GTokenLLMs fully understand graph tokens in the natural-language embedding space?} Motivated by this question, we formalize a unified framework for GTokenLLMs and propose an evaluation pipeline, \textbf{GTEval}, to assess graph-token understanding via instruction transformations at the format and content levels. We conduct extensive experiments on 6 representative GTokenLLMs with GTEval. The primary findings are as follows: (1) Existing GTokenLLMs do not fully understand graph tokens. They exhibit over-sensitivity or over-insensitivity to instruction changes, and rely heavily on text for reasoning; (2) Although graph tokens preserve task-relevant graph information and receive attention across LLM layers, their utilization varies across models and instruction variants; (3) Additional instruction tuning can improve performance on the original and seen instructions, but it does not fully address the challenge of graph-token understanding, calling for further improvement.
\end{abstract}
\section{Introduction}\label{sec: intro}

% Graph无处不在-》GNN建模图数据的成功，但是存在泛化性问题-》LLM的成功促使应用于途任务，吸引大量关注
Graph-structured data, an essential and prevalent form in the
real world, plays a vital role in modeling interactions among objects, such as social networks~\cite{guo2020deep}, e-commerce networks~\cite{zhang2022efraudcom} and so on.
Traditional message-passing graph neural networks (GNNs)~\cite{gcn} are powerful tools for modeling graph structures, but they suffer from a weak multi-task handling capability~\cite{chenllaga}.
Recently, the success of GPT-style LLMs~\cite{llama,gpt} has motivated numerous studies to adapt them to graph learning, particularly for text-attributed graphs (TAGs), where each node is associated with a textual attribute.
The ultimate goal is to build a graph foundation model that generalizes across diverse scenarios~\cite{liu2023towards}. In this setting, LLMs are typically employed as predictors to solve various graph tasks (e.g., node classification), showing promising potential and attracting considerable attention~\cite{chen2024exploring,fatemitalk,konggofa,zhang2025can,wangunigte}.      
\begin{figure}[t]   
    \centering
    \includegraphics[width=1.0\linewidth]{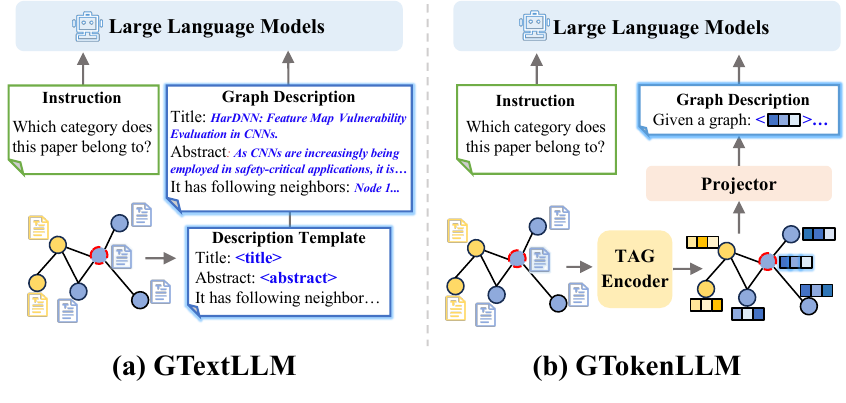}
    \vskip -0.075in
    \caption{Two paradigms for applying LLMs to graph tasks.}
    \label{fig:llm-graph}
    \vskip -0.2in
\end{figure}

A central challenge in this endeavor is aligning graph data with the natural language embedding space so that LLMs can better comprehend graphs. To address this challenge, existing methods primarily follow two paradigms: Graph-Textualizing LLMs (GTextLLMs)~\cite{tan2024walklm,zhao2023graphtext,chen2024exploring,chenlabel} and Graph-Tokenizing LLMs (GTokenLLMs)~\cite{konggofa,chenllaga,tang2024graphgpt}.
GTextLLMs utilize natural language to describe graphs via manually crafted prompt templates, directly aligning them with text, which is then combined with task instructions to query the LLM.
As shown in Figure~\ref{fig:llm-graph}(a), for an academic TAG, node features and topology are formatted into a template description. 
This description includes a paper's title/abstract and its neighbors' titles.
Although straightforward and easy to implement, GTextLLMs struggle to capture complex graph structures and long-range context in natural language, as shown by extensive benchmarks~\cite{huang2023can,wang2024nlgraph}, which raises doubts about their effectiveness.

In contrast, GTokenLLMs compress graph data with complex structures and lengthy text into the LLM input space, producing graph tokens that can be concatenated with natural-language instruction tokens to query the LLM. As shown in Figure~\ref{fig:llm-graph}(b), GTokenLLMs first encode the graph into embeddings using a TAG encoder (e.g., a GNN) to capture topological and textual information, and then transform these embeddings into the LLM space via a learned projector to yield graph tokens. As a result, a GTokenLLM represents the input as ``Given a graph $<$graph$>$'', where ``$<$graph$>$'' is replaced with the resulting graph tokens.
By compressing data into graph tokens, GTokenLLMs effectively mitigate the limitations of GTextLLMs and have been recognized as a mainstream paradigm in recent years. However, such highly compressed graph tokens may become the Achilles' heel of GTokenLLMs, because they are projected from another space and do not correspond to any word in the LLM's vocabulary, making their semantics opaque. Therefore, although GTokenLLMs have been widely recognized, the following question remains largely unknown:
\begin{center}
\emph{Do GTokenLLMs fully understand graph tokens in the natural-language embedding space?}
\end{center}
Understanding this question not only advances the development of reliable and explainable GTokenLLM applications, but also serves as a key touchstone for assessing the soundness and long-term viability of the GTokenLLM paradigm.

% 回答这个问题的挑战
However, this question is challenging to answer. First, the lack of a unified, systematic framework for GTokenLLMs hinders the identification of model failures, preventing an in-depth understanding of the underlying causes.
Second, there is a lack of evaluation tools or pipelines to comprehensively investigate this question. Current evaluations rely on simple, fixed instruction templates for each task, which inherently fail to explore the vast natural-language space.

% 我们做了啥
In this paper, we first formalize a unified framework for GTokenLLMs by characterizing the stages of graph-data transformation, providing a clear and comprehensive view of this paradigm. Second, we propose \textbf{GTEval}, an evaluation pipeline that assesses graph-token understanding in GTokenLLMs through instruction transformations at the format and content levels. Finally, we conduct extensive experiments with GTEval to provide deeper insights into GTokenLLMs. The main observations are as follows:

% The main observations are summarized follow:
$\bullet$ Existing GTokenLLMs do not fully understand graph tokens, exhibiting either over-sensitivity or over-insensitivity to instruction changes, and relying heavily on textual attributes for reasoning. Thus, they struggle to respond effectively to both format- and content-level instruction changes.\\
$\bullet$ Although graph tokens preserve task-relevant textual and structural information in graph data and are attended to across all layers of the LLM, their utilization patterns differ across models and instruction variants.\\
$\bullet$ Extra instruction tuning improves performance on the original and seen instructions, but it does not fully address the challenge of graph-token understanding. Moreover, GTextLLMs underperform GTokenLLMs on the graph tasks despite stronger language understanding.

% $\bullet$ Existing GTokenLLMs do not fully understand graph tokens; instead, they are over-fitted to training task instructions, and heavily rely on text for reasoning. Consequently, they struggle to respond effectively to both format- and content-level instruction changes. \\
% $\bullet$ For format-level changes, GTokenLLMs tend to be over-sensitive to minor modifications, yielding unreliable or even illegal outputs; whereas for content-level changes, they tend to be over-insensitive, struggling to transfer to new tasks and even repeating the original task labels. \\
% $\bullet$ Instruction tuning on a broader set of instruction variants can improve task performance for both the original instruction and some variants; however, it does not fully address the challenge of graph token understanding in GTokenLLMs.

\section{The Unified Framework of GTokenLLMs}\label{sec:framework}
\begin{figure*}[t]
    \centering
    \includegraphics[width=1.0\textwidth]{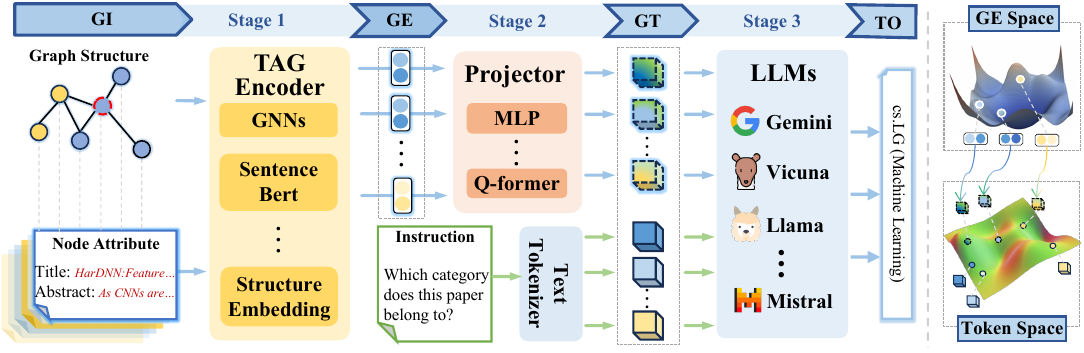}
    \vskip -0.05in
    \caption{The GTokenLLM framework, illustrating a stage-wise transformation of graph data from raw graph inputs (GI) to graph embeddings (GE), projected graph tokens (GT), and finally LLM-generated textual outputs (TO).}
    \vskip -0.2in
    \label{fig:gtokenllm}
\end{figure*}

\subsection{The Formal Definition of GTokenLLM Framework}
We systematically define the formal framework of GTokenLLMs, grounded in the observation that existing GTokenLLMs typically follow a common paradigm: graph data with complex structures and lengthy texts are encoded and aligned to the natural language embedding space, and then used as prefix tokens to query LLMs for downstream tasks. As shown in Figure~\ref{fig:gtokenllm}, we characterize GTokenLLMs by the stages of graph data transformation, from raw graph inputs (Graph Input, GI) to encoded embeddings (Graph Embedding, GE), projected token representations (Graph Token, GT), and finally LLM-generated outputs (Text Output, TO).

\vspace{-7pt}
\paragraph{\textbf{Graph Input (GI).}} As input to GTokenLLMs, the text-attributed graph (TAG) can be viewed as a combination of graph structure and text attributes, where each node is associated with textual information. Formally, a TAG is defined as $\mathcal{G}\!\!=\!\!(\mathcal{V}, \mathcal{E}, \mathcal{T})$, where 
$\mathcal{V}\!=\!\{v_i\}_{i=1}^{\left | \mathcal{V} \right |}, \mathcal{E}\!=\!\{e_i\}_{i=1}^{\left | \mathcal{E} \right |}, \mathcal{T}\!=\!\{t_i\}_{i=1}^{\left | \mathcal{T} \right |}$ denote the node set, edge set, and text set, respectively. The adjacency matrix of $\mathcal{G}$ is denoted as $\mathbf{A}\!\in\!\mathbb{R}^{\left | \mathcal{V} \right | \times \left | \mathcal{V} \right |}$, where $\mathbf{A}_{ij}\!=\!1$ if nodes $v_i$ and $v_j$ are connected, and $\mathbf{A}_{ij}\!=\!0$ otherwise.
Existing GTokenLLMs primarily consider the node classification task on TAGs. Specifically, each node $v_i$ is associated with a textual label $y_i$ indicating its category, and all such labels form the label set $\mathcal{Y}$. The goal is to predict the labels of nodes in the test set $\mathcal{V}_{\text{test}}$.

\vspace{-7pt}
\paragraph{\textbf{Stage 1: GI $\rightarrow$ GE.}}
In the first stage, a TAG encoder is commonly used to compress complex graph structures and lengthy textual attributes into GEs $\mathbf{Z}\!=\!\{\mathbf{z}_i\}_{i=1}^{|\mathcal{V}|}$.
Specifically, an off-the-shelf text encoder (e.g., BERT~\cite{devlin2019bert}) is first used to encode the lengthy textual attributes, yielding the node features $\mathbf{X}\!\!=\!\!\{\mathbf{x}_i\}_{i=1}^{|\mathcal{V}|}$. Subsequently, structural features can be captured in two main ways. One approach is to use explicit structure embeddings derived from $\mathbf{A}$ as additional feature inputs (e.g., Laplacian embedding~\cite{dwivedi2020generalization}), and then fuse them with $\mathbf{X}$ to form $\mathbf{Z}$~\cite{chenllaga, ji2025anchors}. The other approach is to employ a GNN $f_g$, which implicitly captures structural information through message passing, i.e., $\mathbf{Z}\!=\!f_g(\mathbf{X}, \mathbf{A})$~\cite{wang2024llms, konggofa}.

\vspace{-7pt}
\paragraph{\textbf{Stage 2: GE $\rightarrow$ GT.}} 
In this stage, GEs are mapped and aligned to the LLM token space via a learned projector, yielding GTs. Since GEs and language tokens reside in distinct spaces in terms of both feature dimensions and semantics, GEs are not directly compatible with LLMs for querying, unlike textual tokens. To bridge this gap, GTokenLLMs train a projector $f_p$ to map $\mathbf{Z}$ into GTs, i.e., $\mathbf{T} = f_p(\mathbf{Z})$, using graph tasks (e.g., node classification).

\vspace{-7pt}
\paragraph{\textbf{Stage 3: GT $\rightarrow$ TO.}}
In the final stage, GTs are concatenated with other textual content, mainly task instructions, to query the LLM and generate TOs $\hat{y}$. 
This process relies on a manually designed prompt template that describes the graph, where special placeholders are predefined and later replaced by GTs during tokenization. The exact form of the template depends on the prompting strategy, such as where GTs are placed and how structural information is expressed. For example, a prompt template may take the form ``Given a citation graph $<$graph$>$. Which category does this paper belong to?'', where $<$graph$>$ is a placeholder replaced by the GTs of the central node and its neighbors.

In this paper, we investigate 6 representative GTokenLLMs: LLaGA~\cite{chenllaga}, InstructGLM~\cite{instructglm}, GraphGPT~\cite{tang2024graphgpt}, GraphTranslator~\cite{gtr}, TEA-GLM~\cite{wang2024llms}, and GOFA~\cite{konggofa}. We show that these models can be fitted into our proposed framework. Given the limited space, please refer to Appendix~\ref{sec:gtokenllm} for more details.

\section{GTEval: Evaluating the Graph Token Understanding of GTokenLLMs} \label{sec:gteval}
% In this section, we propose GTEval, a pipeline for evaluating whether GTokenLLMs fully understand mixed inputs of graph tokens and language tokens.
\subsection{The Definition of GTEval}
We characterize the understanding of graph tokens at two levels.
At the format level, understanding requires that GTokenLLMs are not only fitted to specific task instructions in the training set, but can also appropriately respond to format changes, such as task paraphrasing.
At the content level, understanding means that GTokenLLMs can generalize to new task instructions with different semantic content, given the same GT inputs.
Therefore, we design four evaluation methods in GTEval by transforming the original task instructions at both the format and content levels, to assess format-level (Rephrasing) and content-level (Reversing, Relabeling, and Randomizing) understanding of GTs in GTokenLLMs.

\vspace{-7pt}
\paragraph{\textbf{Original task instruction.}}
Current evaluations of GTokenLLMs rely on a single, fixed instruction template for each task. As illustrated in Table~\ref{tab:example}, the first box presents the textual features of an example node belonging to the class ``cs.LG (Machine Learning).'' The second box shows the original instruction template used during both training and evaluation. In this template, the black text is fixed, while the blue text is instantiated with GEs and label lists. In practice, since GTokenLLMs output free-form text, \textbf{\textit{Accuracy} (Acc$\uparrow$)} is computed by counting a prediction as correct if it contains the ground-truth label and no other label~\cite{chenllaga}. Formally, accuracy is defined as:
\begin{align}
\text{Accuracy}
&\!=\! \frac{1}{|\mathcal{V}_{\text{test}}|}\!
\sum_{v_i \in \mathcal{V}_{\text{test}}}
\!\!\!\mathbb{I}\Big(
\operatorname{contains}(\hat{y}_i, y_i)
\nonumber \\
&\quad \land\;
\!\forall y \!\in\! \mathcal{Y}\!\setminus\!\{y_i\},\;
\neg \operatorname{contains}(\hat{y}_i, y)
\Big),
\label{eq:accuracy}
\end{align}
where $\hat{y}_i$ denotes the textual output generated by the LLM,
$y_i \! \in \! \mathcal{Y}$ is the ground-truth textual label,
$\operatorname{contains}(\hat{y}_i, y)$ indicates that the output text $\hat{y}_i$ contains the label string $y$,
and $\mathbb{I}(\cdot)$ is the indicator function.

\begin{table*}[t]
\caption{Illustrative examples of instruction types instantiated for the LLaGA in GTEval, with key modifications \underline{underlined}.}
% \caption{Illustrative examples of instruction types in GTEval instantiated for LLaGA, with key modifications \underline{underlined}.}
% \caption{Example of instruction types in GTEval. Key modifications that correspond to the instruction type are \underline{underlined}.}
\vskip -0.175in
\label{tab:example}
\begin{center}
\begin{small}
\setlength{\extrarowheight}{1pt}
\resizebox{1\textwidth}{!}{
\begin{tabular}{p{17cm}}
\toprule
\textbf{Node ID:} 43800 \quad\quad\quad\quad\quad\quad \textbf{Title:} HarDNN: Feature Map Vulnerability Evaluation in CNNs.  \\
\textbf{Abstract:} As CNNs are increasingly being employed in safety-critical applications, it is important that they behave reliably ... This paper presents HarDNN, a software-directed approach ...
 % \\
\quad\quad\quad\quad\quad\quad\quad \textbf{Label: } cs.LG(Machine Learning) \\[2pt]
\hline
\parbox[c][0.4cm][c]{0cm}{}\textbf{\textcolor{brown}{Instruction Type: }}\textbf{Original}\\
\textcolor{olive}{\textbf{Instruction Template: }}Given a node-centered graph: \promptfield{graph}, we need to classify the center node into 40 classes: \promptfield{cs.LG (Machine Learning), cs.AI(Artificial Intelligence), ... , cs.CR (Cryptography and Security)}, please tell me which class the center node belongs to?  \\
\textbf{\textcolor{softgreen}{Expected Output: }}cs.LG (Machine Learning).\quad\quad\quad\quad\quad\quad\quad\quad\quad\quad\quad
\textcolor{softgreen}{\textbf{Possible Output (\ding{51}): }}cs.LG (Machine Learning).  \\
\bottomrule

\parbox[c][0.4cm][c]{0cm}{}\textbf{\textcolor{brown}{Instruction Type: }}\textbf{Rephrasing}\\
\textcolor{olive}{\textbf{Instruction Template: }}\underline{Using} the \underline{node-centric} graph: \promptfield{graph}, \underline{classify} the central node into \underline{one of the 40 given categories}: \promptfield{cs.LG (Machine Learning), cs.AI(Artificial Intelligence), ... , cs.CR(Cryptography and Security)}.  \\
\textbf{\textcolor{softgreen}{Expected Output: }}cs.LG (Machine Learning).\quad\quad\quad\quad\quad\quad\quad\quad\quad\quad\quad
\textcolor{softred}{\textbf{Possible Output (\ding{55}): }}cs.CR (Cryptography and Security). \\
\bottomrule

\parbox[c][0.4cm][c]{0cm}{}\textbf{\textcolor{brown}{Instruction Type: }}\textbf{Relabeling}\\
\textcolor{olive}{\textbf{Instruction Template: }}Using the node-centric graph: \promptfield{graph}, classify the central node into one of the \underline{given categories}: \underline{\promptfield{Artificial Intelligence and Machine Learning, Theoretical Computer Science, ..., Applied Computing and Engineering}}.  \\
\textbf{\textcolor{softgreen}{Expected Output: }}Artificial Intelligence and Machine Learning.\quad\quad\quad\!\!\quad\,
\textcolor{softred}{\textbf{Possible Output (\ding{55}): }}Applied Computing and Engineering.   \\
\bottomrule

\parbox[c][0.4cm][c]{0cm}{}\textbf{\textcolor{brown}{Instruction Type: }}\textbf{Reversing}\\
\textcolor{olive}{\textbf{Instruction Template: }}Using the node-centric graph: \promptfield{graph}, classify the central node into the \underline{least probable} of the 40 given categories: \promptfield{cs.LG (Machine Learning), cs.AI (Artificial Intelligence), ... , cs.CR (Cryptography and Security)}.  \\
\textbf{\textcolor{softgreen}{Expected Output: }} cs.CR (Cryptography and Security).\quad\quad\quad\quad\!\!\!\quad\quad\quad\ \,
\textcolor{softred}{\textbf{Possible Output (\ding{55}): }}cs.LG (Machine Learning). \\
\bottomrule

\parbox[c][0.4cm][c]{0cm}{}\textbf{\textcolor{brown}{Instruction Type: }}\textbf{Randomizing}\\
\textcolor{olive}{\textbf{Instruction Template: }}Using the node-centric graph: \promptfield{graph}, \underline{the ducks are planning to organize a concert in the park}: \promptfield{cs.LG (Machine Learning), cs.AI(Artificial Intelligence), ..., cs.CR  (Cryptography and Security)}.  \\
\textbf{\textcolor{softgreen}{Expected Output: }}Any answer that is not related to node classification.\quad 
\textcolor{softred}{\textbf{Possible Output (\ding{55}): }}cs.LG (Machine Learning). \\
\bottomrule

\end{tabular}
}
\end{small}
\end{center}
\vskip -0.25in
\end{table*}
\vspace{-7pt}
\paragraph{\textbf{Format-level modification.}}
A basic requirement for GTokenLLMs is to produce consistent responses for the same GTs when the task format changes but the underlying semantics remain the same. To evaluate this property, we define \textbf{Rephrasing} as paraphrasing the instruction while keeping the task itself (e.g., node classification) unchanged. Minor edits, such as synonym replacement or word reordering, are allowed in task-irrelevant parts.
As shown in the third box of Table~\ref{tab:example}, terms such as ``Given,'' ``node-centered,'' and ``classify \dots into 40 classes'' are rephrased as ``Using,'' ``node-centric,'' and ``classify \dots into one of the 40 given categories,'' respectively. A GTokenLLM that fails to properly understand GTs at the format level may produce inconsistent predictions under such rephrased instructions.
Notably, since the task semantics remain unchanged, we use the accuracy defined in Equation~\ref{eq:accuracy} as the evaluation metric.

\vspace{-5pt}
\paragraph{Content-level modification.} 
A higher expectation is that GTokenLLMs respond correctly to changes in task content given the same GT inputs. Here, we define three modification methods. ``Relabeling'' and ``Reversing'' evaluate the model's adaptability to changes in label space and semantics; in these settings, the model should adapt to the new instructions with GTs, as the tasks are similar but differ in labels or structure. 
``Randomizing'' assesses the worst-case scenario by introducing random task instructions. If models are overfitted to training tasks, they may still produce answers for the original task despite the random instructions.

$\bullet$ \textbf{Relabeling:} 
This instruction replaces the original label space with a human-defined one that is human-readable and reasonable. As illustrated in the fourth box of Table~\ref{tab:example}, the original label space ``cs.LG (Machine Learning), cs.AI (Artificial Intelligence), ..., cs.CR (Cryptography and Security)'' is replaced with ``Artificial Intelligence and Machine Learning, Information Systems and Data Management, ..., Applied Computing and Engineering''. This transition is achieved by summarizing the original labels. For instance, ``cs.LG (Machine Learning)'' becomes ``Artificial Intelligence and Machine Learning''. 
If GTokenLLMs fail to capture the intrinsic semantics of the original labels, they may be unable to produce the correct relabeled outputs, leading to incorrect or even invalid predictions. Formally, we further define the \textbf{\textit{Relabel} ($\uparrow$)} metric to evaluate models:\\
\vspace{-15pt}
\begin{align}
\text{Relabel}
&\!=\! \frac{1}{|\mathcal{V}_{\text{test}}|}\!
\sum_{v_i \in \mathcal{V}_{\text{test}}}
\!\!\!\mathbb{I}\Big(
\operatorname{contains}(\hat{y}_i, y_i^{\text{rel}})
\nonumber \\
&\quad \land\;
\!\forall y \!\in\! \mathcal{Y}^{\text{rel}}\!\setminus\!\{y_i^{\text{rel}}\},\;
\neg \operatorname{contains}(\hat{y}_i, y)
\nonumber \\
&\quad \land\;
\!\forall y \!\in\! \mathcal{Y},\;
\neg \operatorname{contains}(\hat{y}_i, y)
\Big),
\label{eq:relabel}
\end{align}
% \vspace{-5pt}
where $y_i^{\text{rel}} \!\in\! \mathcal{Y}^{\text{rel}}$ is the relabeled ground-truth label of node $v_i$, and $\mathcal{Y}^{\text{rel}}$ denotes the relabeled label set.

$\bullet$ \textbf{Reversing:} 
Unlike Relabeling, this setting reverses the objective of the original task, forcing the model to break its default preference.
As illustrated in the fifth box of Table~\ref{tab:example}, the phrase ``least probable'' is added to require the model to identify the most unrelated category instead of the most relevant one.
If GTokenLLMs fail to understand the semantics of graph tokens, they may be stuck at memorizing label names rather than performing discriminative reasoning, and thus still output the original label. We further define the \textbf{\textit{Reverse} ($\uparrow$)} metric to evaluate this capability as follows:
\begin{align}
\mathcal{V}_{\mathrm{corr}}
&\!=\!\Big\{v_i \!\in\! \mathcal{V}_{\text{test}}
\ \Big| \ 
\operatorname{contains}(\hat{y}_i^{\text{ori}}, y_i)
\nonumber \\
&\quad \land\;
\!\forall y \!\in\! \mathcal{Y}\!\setminus\!\{y_i\},\;
\neg \operatorname{contains}(\hat{y}_i^{\text{ori}}, y)
\Big\},
\label{eq:vcorr}
\\
\text{Reverse}
&\!=\! 1 - \frac{1}{|\mathcal{V}_{\mathrm{corr}}|}\!
\sum_{v_i \in \mathcal{V}_{\mathrm{corr}}}
\!\!\!\mathbb{I}\Big(
\operatorname{contains}(\hat{y}_i^{\text{rev}}, y_i)
\nonumber \\
&\quad \land\;
\!\forall y \!\in\! \mathcal{Y}\!\setminus\!\{y_i\},\;
\neg \operatorname{contains}(\hat{y}_i^{\text{rev}}, y)
\Big),
\label{eq:rev}
\end{align}
where $\hat{y}_i^{\text{ori}}$ and $\hat{y}_i^{\text{rev}}$ are the textual outputs generated by the GTokenLLM under the original and reversing instructions for node $v_i$, respectively.

\begin{table*}[t]
\centering
% \caption{Evaluating whether GTokenLLMs understand graph tokens, where ($\Delta$) compared to the Original and $\varnothing$ denotes illegal output.}
% \caption{Evaluation results of GTokenLLMs on GTEval. $\Delta$ denotes the performance change relative to the original and $\varnothing$ indicates illegal outputs.}
\caption{Evaluation results of GTokenLLMs on GTEval. $\Delta$ denotes the change from the original, and $\varnothing$ indicates illegal outputs.}

\label{tab:main}
\vskip -0.075in
% \label{tab:main_results}
% \setlength{\tabcolsep}{4pt}
% \renewcommand{\arraystretch}{0.9}
\setlength{\extrarowheight}{-0.55pt}
\large
\resizebox{\textwidth}{!}{
\begin{tabular}{ccccl|ccc}
\toprule
\multirow{2}{*}{\textbf{Model}} & \multirow{2}{*}{\textbf{Dataset}} 
& \multicolumn{3}{c|}{\textbf{Format-Level}} 
& \multicolumn{3}{c}{\textbf{Content-Level}} \\
\cmidrule(lr){3-5}\cmidrule(lr){6-8}
& & \textbf{Ori. Acc ($\uparrow$)} & \textbf{Rep. Acc ($\uparrow$)} & \multicolumn{1}{c|}{\textbf{$\Delta$}}
& \textbf{Relabel ($\uparrow$)} & \textbf{Reverse ($\uparrow$)} & \textbf{Random ($\uparrow$)} \\
\midrule

\multirow{5}{*}{\textbf{LLaGA}} 
& Cora     & 87.45 & 86.44$\pm$0.93 & \red{$\downarrow$1.16\%} & 5.74$\pm$3.90  & 1.73$\pm$0.95  & 0.85$\pm$0.54 \\
& Pubmed   & 92.13 & 92.39$\pm$0.38 & \green{$\uparrow$0.28\%} & 40.30$\pm$7.89 & 1.48$\pm$0.55  & 1.93$\pm$2.68 \\
& Arxiv    & 74.36 & 69.55$\pm$2.34 & \red{$\downarrow$6.47\%} & 1.19$\pm$0.49  & 4.76$\pm$0.75  & 5.59$\pm$1.65 \\
& Products & 83.22 & 81.79$\pm$1.09 & \red{$\downarrow$1.72\%} & 6.84$\pm$0.85  & 2.84$\pm$1.39  & 3.76$\pm$0.30 \\ \cmidrule(lr){2-8}
& Overall  & 84.29 & 81.54          & \red{$\downarrow$3.26\%} & 13.52          & 2.70           & 3.03 \\
\midrule

\multirow{4}{*}{\textbf{InstructGLM}} 
& Cora    & 78.59 & 39.24$\pm$19.71 & \red{$\downarrow$50.07\%} & $\varnothing$ & 15.81$\pm$15.25 & $\varnothing$ \\
& Pubmed  & 91.87 & 55.05$\pm$22.67 & \red{$\downarrow$40.08\%} & $\varnothing$ & 9.46$\pm$10.66  & $\varnothing$ \\
& Arxiv   & 76.85 & 74.96$\pm$1.32  & \red{$\downarrow$2.45\%}  & $\varnothing$ & 18.63$\pm$8.33  & $\varnothing$ \\  \cmidrule(lr){2-8}
& Overall & 82.44 & 56.42           & \red{$\downarrow$31.56\%} & $\varnothing$ & 14.63           & $\varnothing$ \\
\midrule

\multirow{4}{*}{\textbf{GraphGPT}} 
& Cora-70 & 18.66 & 18.40$\pm$1.25 & \red{$\downarrow$1.41\%}  & 30.26$\pm$1.04 & 74.51$\pm$10.62 & 89.83$\pm$1.93 \\
& Pubmed  & 80.71 & 74.21$\pm$6.39 & \red{$\downarrow$8.05\%}  & 57.74$\pm$1.90 & 40.63$\pm$18.07 & 58.27$\pm$4.19 \\
& Arxiv   & 62.50 & 39.80$\pm$12.26& \red{$\downarrow$36.32\%} & 21.46$\pm$1.37 & 27.68$\pm$17.84 & 99.00$\pm$0.27 \\  \cmidrule(lr){2-8}
& Overall & 53.96 & 44.14          & \red{$\downarrow$18.20\%} & 36.49          & 47.61           & 82.37 \\
\midrule

\multirow{4}{*}{\textbf{TEA-GLM}} 
& Cora-70 & 12.10 & 10.06$\pm$1.28 & \red{$\downarrow$16.84\%} & 21.46$\pm$5.99  & 34.84$\pm$14.89 & 76.71$\pm$3.66 \\
& Pubmed  & 79.54 & 59.09$\pm$19.40& \red{$\downarrow$25.71\%} & 60.33$\pm$24.66 & 60.97$\pm$25.76 & 40.05$\pm$4.79 \\
& History & 50.76 & 38.05$\pm$7.36 & \red{$\downarrow$25.03\%} & 24.05$\pm$5.93  & 70.75$\pm$16.66 & 64.16$\pm$5.03 \\  \cmidrule(lr){2-8}
& Overall & 47.47 & 35.73          & \red{$\downarrow$24.73\%} & 35.28           & 55.52           & 60.31 \\
\midrule

\multirow{3}{*}{\textbf{GOFA}} 
& Cora    & 68.91 & 67.70$\pm$0.79 & \red{$\downarrow$1.75\%} & 72.49$\pm$25.48 & 2.03$\pm$0.87 & 4.48$\pm$4.86 \\
& WikiCS  & 71.06 & 69.77$\pm$0.54 & \red{$\downarrow$1.81\%} & 76.63$\pm$0.84  & 3.97$\pm$2.67 & 3.77$\pm$1.30 \\  \cmidrule(lr){2-8}
& Overall & 69.98 & 68.74          & \red{$\downarrow$1.78\%} & 74.56           & 3.00          & 4.13 \\
\midrule

\multirow{1}{*}{\textbf{GraphTranslator}} 
& Arxiv   & 26.59 & 16.02$\pm$2.87 & \red{$\downarrow$39.73\%} & 20.90$\pm$3.65 & 70.71$\pm$9.37 & 91.11$\pm$8.63 \\
\bottomrule
\end{tabular}
}
\vskip -0.175in
\end{table*}

$\bullet$ \textbf{Randomizing:} This modification considers an extreme case where the original task instruction is replaced with an irrelevant sentence, to test whether a GTokenLLM can distinguish valid task instructions from unrelated text. Under this setting, the model is expected to refrain from producing task-specific outputs. However, a GTokenLLM that fails to distinguish the task itself may still respond as if performing the original task. As illustrated in the sixth box of Table~\ref{tab:example}, when the node classification instruction is replaced by an irrelevant sentence such as ``the ducks are planning to organize a concert in the park,'' the model may still output the original classification result, e.g., ``cs.LG (Machine Learning).'' Here, we further define the \textbf{\textit{Random} ($\uparrow$)} metric:
\begin{align}
\text{Random}
&\!=\! \frac{1}{|\mathcal{V}_{\text{test}}|}\!\!
\sum_{v_i \in \mathcal{V}_{\text{test}}}
\!\!\mathbb{I}\Big(
\forall y \!\in\! \mathcal{Y},\;
\neg \operatorname{contains}(\hat{y}_i, y)
\Big).
\label{eq:random}
\end{align}
\vspace{-16pt}
\subsection{Evaluation Settings of GTEval}
\vspace{-1pt}
In this paper, we evaluate 6 representative GTokenLLMs, including LLaGA~\cite{chenllaga}, InstructGLM~\cite{instructglm}, GraphGPT~\cite{tang2024graphgpt}, GraphTranslator~\cite{gtr}, TEA-GLM~\cite{wang2024llms}, and GOFA~\cite{konggofa}. To more faithfully reflect the models' performance in practice, we directly use the released checkpoint weights and evaluate each model on its corresponding test set, without additional training or fine-tuning. The dataset statistics used by different models are summarized in Table~\ref{tab:dataset_stats}.
To construct test instructions efficiently and reliably, we design four instruction modification templates with the assistance of ChatGPT, followed by manual verification by domain experts. For each dataset and model, we generate ten variants per template to mitigate randomness introduced by wording variations. These instruction variants are then instantiated with the test data for evaluation. We report the mean and standard deviation of the results across different instruction modification templates. More experiment details are provided in Appendix~\ref{sec: exp_setting_detail}.

\section{Experimental Results on GTEval}
In this section, we conduct extensive experiments with GTEval to explore whether GTokenLLMs fully understand GTs. The results indicate that their capabilities in this regard remain at a preliminary stage. To better understand the underlying reasons, we further perform a series of analytical experiments that inspect each stage of the GTokenLLM framework defined in Section~\ref{sec:framework}. Finally, we explore two practical directions to alleviate this limitation: instruction tuning on template variants and directly adopting GTextLLMs.

\subsection{Overall Performance on GTEval}
Table~\ref{tab:main} reports the results of 6 representative GTokenLLMs evaluated with GTEval. We have the following observations:

\begin{figure*}[tp]
    \centering
    \begin{minipage}[b]{0.52\textwidth}
        \centering
        \includegraphics[width=\linewidth]{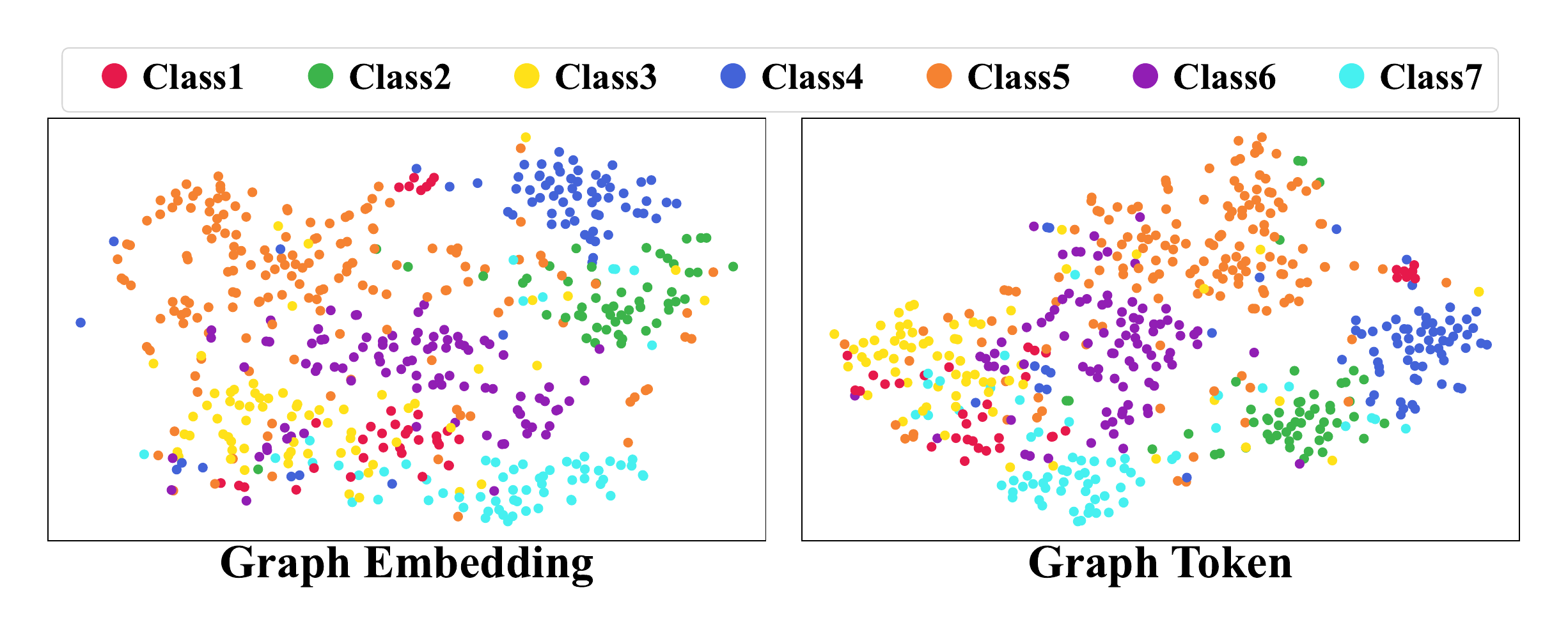}
        \vskip -0.15in
        \caption{t-SNE visualization of GEs and GTs on Cora.}
        \vskip -0.15in
        \label{fig:cora_tsne}
    \end{minipage}
    \hfill
    \begin{minipage}[b]{0.45\textwidth}
        \centering
        \includegraphics[width=0.85\linewidth]{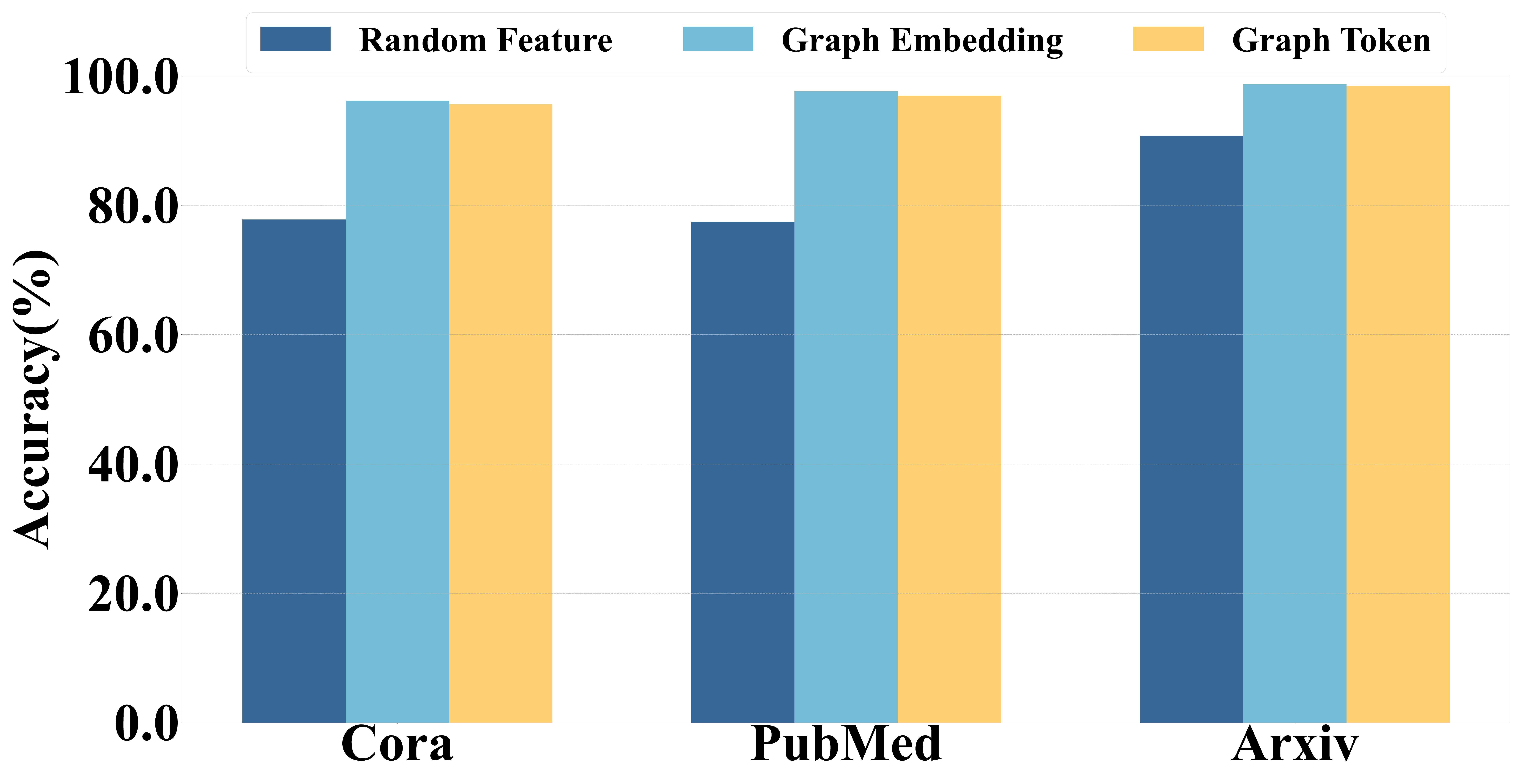}
        \vskip -0.11in
        \caption{Link prediction with GEs, GTs, and random features.}
        \vskip -0.15in
        \label{fig:linkpred}
    \end{minipage}
\end{figure*}
\textbf{\underline{Observation 1:} GTokenLLMs fail to exhibit consistent understanding of GTs across different instruction formulations, struggling to effectively respond to both format- and content-level changes.} No existing GTokenLLM performs well across all instruction variants.
Specifically, these models achieve around 70\%--90\% accuracy under the original instructions. For format changes, Rephrasing results in much lower accuracy, with gaps reaching up to 50\%. For content changes, the performance of most methods is below 50\% and even degrades to under 10\%. InstructGLM even produces invalid outputs such as blank responses or ``yes''. Overall, these results indicate that current GTokenLLMs still exhibit a clearly limited understanding of graph tokens.

\textbf{\underline{Observation 2}: GTokenLLMs exhibit over-insensitivity or over-sensitivity behavior under instruction changes.} 
For example, both LLaGA and GOFA remain stable under the format-level change (Rephrasing), but achieve performance below 6\% under content-level changes (Reversing and Randomizing). This suggests that the models may be severely over-fitted to the training tasks, showing ``over-insensitivity'' to natural language instructions.
In contrast, GraphGPT, TEA-GLM, and GraphTranslator exhibit ``over-sensitivity'' to format-level changes, with overall accuracy variations of approximately 18\%--40\% under Rephrasing, while showing comparatively better results on content-level changes. InstructGLM exhibits over-sensitivity to both format-level and content-level changes, with low accuracy on format changes and invalid outputs on content changes.

\textbf{Summary:} Existing GTokenLLMs exhibit only a preliminary understanding of GTs, which remains insufficient for full comprehension.
We call for future evaluations not to be restricted to a single fixed task instruction, but to consider diverse task formulations for more reliable assessments.

\begin{table}[t]
\centering
% \caption{Model performance before and after graph structure attacks (PRBCD).}
\caption{Model performance under the original and PRBCD structure attack, where $\Delta$ denotes the change from the original.}
\vskip -0.075in
\label{tab:topo_attack}
% \vspace{-5pt}
% \setlength{\tabcolsep}{8pt}
\renewcommand{\arraystretch}{0.85}
\resizebox{\linewidth}{!}{
\begin{tabular}{cccc l}
\toprule
\textbf{Model} & \textbf{Dataset} & \textbf{Original} & \textbf{PRBCD} & \multicolumn{1}{c}{\textbf{$\Delta$}} \\
\midrule

\multirow{4}{*}{\textbf{GCN}}
& Cora    & 89.85 & 23.43 & \red{$\downarrow$73.92\%} \\
& Cora-70 & 19.15 & 3.03  & \red{$\downarrow$84.18\%} \\
& Pubmed  & 80.88 & 23.58 & \red{$\downarrow$70.85\%} \\
\cmidrule(lr){2-5}
& Overall & 63.30 & 16.70 & \red{$\downarrow$73.65\%} \\
\midrule

\multirow{3}{*}{\textbf{LLaGA}}
& Cora    & 87.45 & 48.71 & \red{$\downarrow$44.30\%} \\
& Pubmed  & 92.13 & 60.29 & \red{$\downarrow$34.56\%} \\
\cmidrule(lr){2-5}
& Overall & 89.79 & 54.40 & \red{$\downarrow$39.30\%} \\
\midrule

\multirow{3}{*}{\textbf{GraphGPT}}
& Cora-70 & 18.66 & 20.99 & \green{$\uparrow$12.49\%} \\
& Pubmed  & 80.71 & 80.46 & \red{$\downarrow$0.31\%} \\
\cmidrule(lr){2-5}
& Overall & 49.69 & 50.73 & \green{$\uparrow$2.09\%} \\
\midrule

\multirow{3}{*}{\textbf{TEA-GLM}}
& Cora-70 & 12.10 & 12.39 & \green{$\uparrow$2.40\%} \\
& Pubmed  & 79.54 & 79.21 & \red{$\downarrow$0.41\%} \\
\cmidrule(lr){2-5}
& Overall & 45.82 & 45.80 & \red{$\downarrow$0.04\%} \\
\bottomrule
\end{tabular}
}
% \vspace{-20pt}
\vskip -0.15in
\end{table}

\subsection{Deeper Insights into Each Stage of GTokenLLMs}
To dive into understanding the underlying reasons, we conduct a stage-wise analysis of GTokenLLMs, following the transformation from GIs to GEs, GTs, and finally TOs.
\subsubsection{Insights from ``GI-to-GE" Stage} \label{sec:gigt}
First, to examine how GIs (including textual attributes and graph structure) affect GTokenLLM inference, we apply PRBCD~\cite{geisler2021robustness} to attack graph topology. The perturbation budget for each node is its degree, enabling effective structural disruption. Table~\ref{tab:topo_attack} reports the accuracy of vanilla GCN~\cite{gcn}, LLaGA, GraphGPT, and TEA-GLM. Notably, LLaGA relies solely on GTs during inference, whereas GraphGPT and TEA-GLM additionally incorporate the textual attributes of the central node.

\textbf{\underline{Observation 3:} GTokenLLMs lack an effective understanding of GTs and heavily rely on textual attributes for reasoning.}
As a structure-dependent baseline, GCN suffers a severe performance drop on all datasets (e.g., 73.65\% overall).
In contrast, although LLaGA also degrades substantially (39.30\% overall), it shows ``over-insensitivity'' to content-level modifications (Table~\ref{tab:main}), suggesting that it may mainly memorize a direct mapping from GTs to labels rather than understand them. Meanwhile, GraphGPT and TEA-GLM remain almost unchanged and even slightly improve in some cases, suggesting that their reasoning is mainly based on the input node text rather than GT understanding.

\begin{table*}[t]
\centering
\caption{The performance of instruction-tuned GTokenLLMs with instruction variants Rephrasing and Randomizing. ``Imp.'' denotes the improvement over the corresponding result in Table~\ref{tab:main} (i.e., without extra instruction tuning).}
\label{tab:it}
\vskip -0.075in
\setlength{\tabcolsep}{3pt}
\renewcommand{\arraystretch}{1.5}
\Huge
\resizebox{\textwidth}{!}{
\begin{tabular}{cccl|cl|cl|cl|cl}
\toprule
\multirow{2}{*}{\textbf{Model}} 
& \multirow{2}{*}{\textbf{Dataset}} 
& \multicolumn{4}{c|}{\textbf{Format-Level}}
& \multicolumn{6}{c}{\textbf{Content-Level}} \\
\cmidrule(lr){3-6}\cmidrule(lr){7-12}
& 
& \textbf{Ori. Acc ($\uparrow$)} & \multicolumn{1}{c|}{\textbf{Imp.}} 
& \textbf{Rep. Acc ($\uparrow$)} & \multicolumn{1}{c|}{\textbf{Imp.}}
& \textbf{Relabel ($\uparrow$)} & \multicolumn{1}{c|}{\textbf{Imp.}}
& \textbf{Reverse ($\uparrow$)} & \multicolumn{1}{c|}{\textbf{Imp.}}
& \textbf{Random ($\uparrow$)} & \multicolumn{1}{c}{\textbf{Imp.}} \\
\midrule

\multirow{4}{*}{\textbf{LLaGA}}
& Cora   & 87.64 & \green{$\uparrow$0.22\%} & 87.36$\pm$0.46 & \green{$\uparrow$1.06\%} & 90.54$\pm$0.60 & \green{$\uparrow$1477.35\%} & 1.14$\pm$0.53 & \red{$\downarrow$34.10\%} & 0.00$\pm$0.00 & \red{$\downarrow$100.00\%} \\
& Pubmed & 95.18 & \green{$\uparrow$3.31\%} & 94.85$\pm$0.24 & \green{$\uparrow$2.66\%} & 96.29$\pm$0.30 & \green{$\uparrow$138.93\%}  & 0.56$\pm$0.29 & \red{$\downarrow$62.16\%} & 0.00$\pm$0.00 & \red{$\downarrow$100.00\%} \\
& Arxiv  & 75.61 & \green{$\uparrow$1.68\%} & 75.46$\pm$0.12 & \green{$\uparrow$8.50\%} & 82.58$\pm$0.23 & \green{$\uparrow$6839.50\%} & 2.99$\pm$0.21 & \red{$\downarrow$37.18\%} & 0.00$\pm$0.00 & \red{$\downarrow$100.00\%} \\
\cmidrule(lr){2-12}
& Overall & 86.14 & \green{$\uparrow$1.76\%} & 85.89 & \green{$\uparrow$3.74\%} & 89.89 & \green{$\uparrow$470.42\%} & 1.56 & \red{$\downarrow$41.15\%} & 0.00 & \red{$\downarrow$100.00\%} \\
\midrule

\multirow{4}{*}{\textbf{GraphGPT}}
& Cora-70 & 20.41 & \green{$\uparrow$9.38\%}  & 18.92$\pm$1.23 & \green{$\uparrow$2.83\%}  & 35.92$\pm$1.84 & \green{$\uparrow$18.70\%} & 59.20$\pm$7.33 & \red{$\downarrow$20.55\%} & 83.24$\pm$1.73 & \red{$\downarrow$7.34\%} \\
& Pubmed  & 89.85 & \green{$\uparrow$11.32\%} & 88.76$\pm$1.37 & \green{$\uparrow$19.61\%} & 91.52$\pm$0.72 & \green{$\uparrow$58.50\%} & 19.60$\pm$16.37 & \red{$\downarrow$51.76\%} & 16.40$\pm$7.03 & \red{$\downarrow$71.86\%} \\
& Arxiv   & 66.40 & \green{$\uparrow$6.24\%}  & 67.19$\pm$0.76 & \green{$\uparrow$68.82\%} & 46.24$\pm$1.02 & \green{$\uparrow$115.47\%} & 4.41$\pm$0.65 & \red{$\downarrow$84.07\%} & 99.52$\pm$0.16 & \green{$\uparrow$0.53\%} \\
\cmidrule(lr){2-12}
& Overall & 58.89 & \green{$\uparrow$9.14\%}  & 58.29 & \green{$\uparrow$32.07\%} & 57.89 & \green{$\uparrow$58.67\%} & 27.74 & \red{$\downarrow$41.74\%} & 66.39 & \red{$\downarrow$19.40\%} \\
\bottomrule
\end{tabular}
}
% \vspace{-5pt}
\vskip -0.15in
\end{table*}

\begin{figure*}[tp]
    \centering
    \includegraphics[width=1\linewidth]{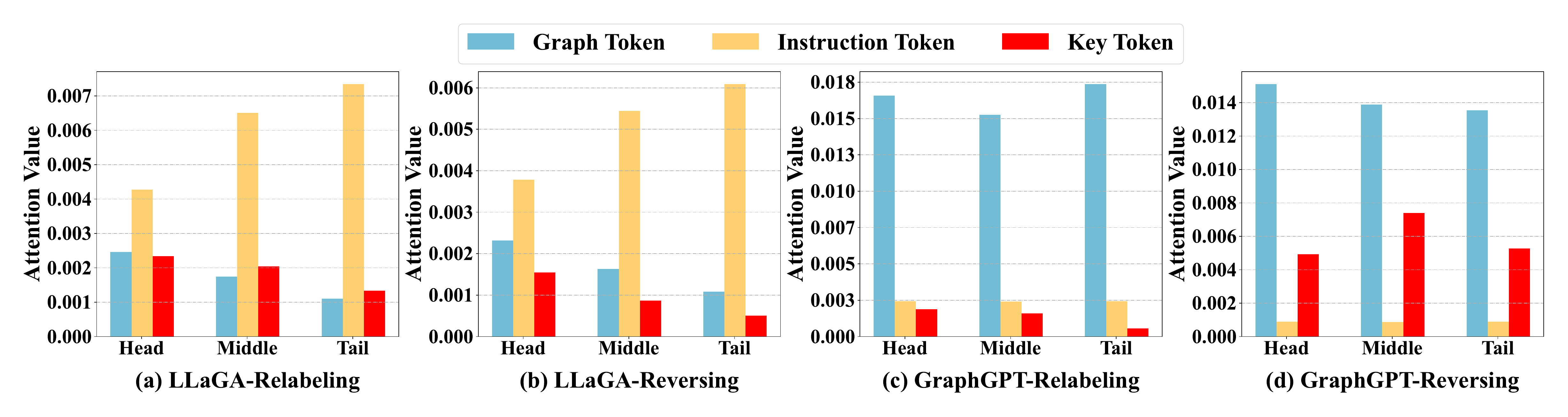}
    \vskip -0.1in
    \caption{Attention distribution of GTokenLLMs under Relabeling and Reversing instruction variants on Cora.}
    \vskip -0.175in
    \label{fig:attn-var}
\end{figure*}
\subsubsection{Insights from ``GE-to-GT" Stage} \label{sec:gegt}
To validate whether GTs retain sufficient textual information, we visualize the embedding spaces of GEs and GTs in LLaGA using t-SNE (Figure~\ref{fig:cora_tsne}). To assess structural information, we train separate GCNs for link prediction using GEs, GTs, and randomly initialized features (Figure~\ref{fig:linkpred}).

\textbf{\underline{Observation 4:} GTs preserve task-relevant textual and structural information in graph data.}
Figure~\ref{fig:cora_tsne} suggests that GTs retain textual signals: nodes with the same label cluster more tightly, and the class boundaries are clearer than those formed by GEs, indicating that supervision from downstream tasks further shapes GTs toward community-level textual cues.
Figure~\ref{fig:linkpred} suggests that GTs also retain structural signals. In link prediction, a GCN trained on GTs performs comparably to one trained on GEs, and both outperform randomly initialized features, implying that edge-level information is largely preserved. The small gap may reflect that GTs encode classification-oriented signals that are not fully aligned with link prediction.
Overall, GTs retain both textual and structural information, especially that related to downstream tasks.

\begin{table*}[t]
\centering
\caption{Evaluation results of the GTextLLM~\cite{chen2024exploring} on GTEval. $\Delta$ denotes the change from the original.}
\label{tab:gtext-res}
% \vskip -0.05in
\vskip -0.075in
\renewcommand{\arraystretch}{0.8}
\resizebox{\textwidth}{!}{
\begin{tabular}{cccc l|ccc}
\toprule
\multirow{2}{*}{\textbf{Model}} & \multirow{2}{*}{\textbf{Dataset}} 
& \multicolumn{3}{c|}{\textbf{Format-Level}} 
& \multicolumn{3}{c}{\textbf{Content-Level}} \\
\cmidrule(lr){3-5}\cmidrule(lr){6-8}
& & \textbf{Ori. Acc ($\uparrow$)} & \textbf{Rep. Acc ($\uparrow$)} 
& \multicolumn{1}{c|}{\textbf{$\Delta$}}
& \textbf{Relabel ($\uparrow$)} & \textbf{Reverse ($\uparrow$)} & \textbf{Random ($\uparrow$)} \\
\midrule

\multirow{4}{*}{\textbf{Zero-Shot}}
& Cora   & 67.15 & 68.41$\pm$0.91 & \green{$\uparrow$1.88\%}  & 58.76$\pm$2.69  & 92.92$\pm$10.89 & 97.40$\pm$1.12 \\
& Pubmed & 78.93 & 82.28$\pm$3.11 & \green{$\uparrow$4.24\%}  & 74.90$\pm$16.25 & 84.03$\pm$17.36 & 99.44$\pm$0.36 \\
& Arxiv  & 54.20 & 52.62$\pm$2.27 & \red{$\downarrow$2.92\%}   & 47.12$\pm$1.36  & 91.86$\pm$15.89 & 95.50$\pm$1.50 \\
\cmidrule(lr){2-8}
& Overall& 66.76 & 67.77          & \green{$\uparrow$1.51\%}  & 60.26           & 89.60           & 97.45 \\
\midrule

\multirow{4}{*}{\parbox[c]{2cm}{\textbf{\centering 1-Hop\\Title+Label}}}
& Cora   & 69.37 & 80.83$\pm$1.71 & \green{$\uparrow$16.52\%} & 80.66$\pm$1.52  & 85.15$\pm$9.13  & 91.16$\pm$5.41 \\
& Pubmed & 88.07 & 86.60$\pm$4.43 & \red{$\downarrow$1.67\%}   & 70.65$\pm$4.92  & 89.14$\pm$3.87  & 100.00$\pm$0.00 \\
& Arxiv  & 47.40 & 60.25$\pm$1.95 & \green{$\uparrow$27.11\%} & 59.82$\pm$3.05  & 93.87$\pm$8.71  & 97.44$\pm$3.92 \\
\cmidrule(lr){2-8}
& Overall& 68.28 & 75.89          & \green{$\uparrow$11.15\%} & 70.38           & 89.39           & 96.20 \\
\bottomrule
\end{tabular}
}
% \vspace{-10pt}
% \vskip -0.05in
\end{table*}

\begin{table*}[t]
\centering
\caption{GTokenLLM with GPT-4o compared to GPT-3.5 on Cora. $\Delta$ denotes the change from the original.}
\label{tab:gpt4}
% \vspace{-5pt}
\vskip -0.075in
    % \vskip -0.15in
\setlength{\tabcolsep}{10pt}
\renewcommand{\arraystretch}{0.8}
\resizebox{\linewidth}{!}{
\begin{tabular}{ccc l|ccc}
\toprule
\multirow{2}{*}{\textbf{Model}}
& \multicolumn{3}{c|}{\textbf{Format-Level}}
& \multicolumn{3}{c}{\textbf{Content-Level}} \\
\cmidrule(lr){2-4}\cmidrule(lr){5-7}
& \textbf{Ori. Acc ($\uparrow$)} & \textbf{Rep. Acc ($\uparrow$)}
& \multicolumn{1}{c|}{\textbf{$\Delta$}}
& \textbf{Relabel ($\uparrow$)} & \textbf{Reverse ($\uparrow$)} & \textbf{Random ($\uparrow$)} \\
\midrule

\textbf{GPT-3.5} & 67.15 & 68.41$\pm$0.91 & \green{$\uparrow$1.88\%} & 58.76$\pm$2.69 & 92.92$\pm$10.89 & 97.40$\pm$1.12 \\
\textbf{GPT-4o}   & 70.11 & 69.00$\pm$1.11 & \red{$\downarrow$1.58\%} & 64.66$\pm$1.84 & 93.16$\pm$5.42  & 94.39$\pm$1.20 \\
\bottomrule
\end{tabular}
}
\vskip -0.15in
\end{table*}

% \begin{table*}[t]
%     \caption{GTokenLLM with GPT-4o compared to GPT-3.5}
%     \label{tab:gpt4}
%     % \vskip -0.15in
%     \centering
%     \begin{tabular}{|l|l|l|l|l|l|l|}
%         \hline 
%         Model & Original ACC & Rephrase ACC & $\Delta$ & $Randomize\ ACC_{rel}$ & $Relabel\ ACC_{ran}$ & $Reverse\ ACC_{rev}$ \\
%         \hline
%         GPT-3.5 & 67.15 & $68.41\pm0.91$ & +1.88\% & $97.40\pm1.12$ & $58.76\pm2.69$ & $92.92\pm10.89$ \\
%         \hline
%         GPT-4 & 70.11 & $69.00\pm1.11$ & -1.58\% & $94.39\pm1.20$ & $64.66\pm1.84$ & $93.16\pm5.42$ \\
%         \hline
%     \end{tabular}
    
% \end{table*}

\subsubsection{Insights from ``GT-to-TO" Stage} 
To examine whether the LLM attends to graph tokens during inference, we analyze attention weights across layers. Following~\cite{darcet2023vision,dosovitskiy2020image}, we extract the cross-attention matrices between output and input tokens from the first, middle, and last three layers (denoted as \emph{head}, \emph{middle}, and \emph{tail}) and average them over attention heads and layers. We then group the attention weights by token type: graph tokens, instruction tokens, and key tokens. Key tokens are the words that distinguish an instruction variant from the original one (e.g., ``least probable'' in Reversing). Figure~\ref{fig:attn-var} shows the results for LLaGA and GraphGPT under Relabeling and Reversing on the Cora dataset.

\textbf{\underline{Observation 5:} GTs are attended to across all layers of the LLM, with attention patterns differing across models and instruction variants.}
Across all layers, LLaGA and GraphGPT assign non-trivial attention to GTs, suggesting that GTs are not ignored when generating TOs. However, the attention distribution differs by model and instruction type.
Specifically, LLaGA allocates most attention to instruction tokens and the least to key tokens. This pattern may bias the model toward the seen training templates and weaken its adaptation to content-level variants.
In contrast, GraphGPT allocates comparable attention to key tokens under Relabeling and even higher attention under Reversing, aligning with its better adaptation to content-level variants.

\subsection{Is Instruction Tuning a Straightforward Remedy?}
Instruction tuning has been acknowledged as an intuitive solution for generalizing to unseen instructions~\cite{zhang2026instruction,kung2023active}. To this end, we construct additional instruction tuning sets. Since ground-truth are unavailable for Randomizing and Reversing, we only add variants of Rephrasing and Relabeling. For each model, we generate 10 templates per variant and instantiate them using the original training data. This setup keeps the training set size unchanged while increasing instruction diversity during training. We apply it to two representative GTokenLLMs, LLaGA and GraphGPT, and report the results in Table~\ref{tab:it}.

\textbf{\underline{Observation 6:} Extra instruction tuning improves performance under the original task instruction.}
Although we do not introduce additional training data, enriching the instruction templates during tuning leads to higher accuracy under the original task instruction. This gain may come from improved instruction-following capability on GT inputs, resulting in more accurate textual predictions. These results suggest that increasing instruction diversity is a simple and practical enhancement during instruction tuning.

\textbf{\underline{Observation 7:} Variant-based instruction tuning improves performance on seen variants but does not generalize to unseen ones.}
The tuned models consistently outperform the original models on the two seen variant types, indicating that instruction tuning is effective when the instruction patterns are covered during training. However, performance shows little improvement and can even degrade on the two unseen types (Randomizing and Reversing), indicating that instruction tuning cannot fundamentally address the deficiency of GTokenLLMs in understanding GTs. This observation also encourages researchers to strive for GTokenLLMs with stronger graph token understanding.

\subsection{Is GTextLLM an Alternative to GTokenLLMs?}
Unlike GTokenLLMs, which learn to align graph tokens with textual tokens, GTextLLM describes graphs in natural language via manually crafted prompt templates, directly aligning graph inputs with text.
For a comprehensive comparison, we evaluate GTextLLM under GTEval and report the results in Table~\ref{tab:gtext-res}, using the prompt templates adopted from Chen et al.~\cite{chen2024exploring}. For consistency with prior GTokenLLM studies, we use GPT-3.5 as the default GTextLLM backbone. We also report GPT-4o results on Cora with the zero-shot prompt in Table~\ref{tab:gpt4}.

\textbf{\underline{Observation 8:} GTextLLMs underperform GTokenLLMs on the same graph task, even though their language understanding is better.}
This gap is expected because GTextLLMs solve graph tasks in a zero-shot, prompt-based manner, whereas GTokenLLMs learn specialized graph representations (e.g., GTs) tailored to downstream supervision. As a result, GTokenLLMs retain a clear advantage in task accuracy~\cite{chen2024exploring,chenllaga}. Moreover, GPT-4o achieves higher accuracy under the original prompt than GPT-3.5, but still falls short of GTokenLLMs (87.45\%). Meanwhile, GPT-4o and GPT-3.5 show similar behavior on GTEval. Overall, these results support the same conclusion across backbones: using natural-language graph descriptions avoids the graph-token understanding issue by construction, yet GTextLLMs remain substantially less competitive than specialized GTokenLLMs on graph tasks.

\section{Conclusion}
Recently, GTokenLLMs have emerged as a mainstream paradigm for extending LLMs to TAGs. In this work, we first systematically examine whether these models fully understand graph tokens. By formalizing GTokenLLMs and introducing GTEval, an evaluation pipeline based on format- and content-level modifications, we assess their ability to adapt to task variations. Our results show that existing GTokenLLMs exhibit only a preliminary stage of graph-token understanding and remain far from reliable semantic comprehension. We hope this study encourages more careful evaluation and deeper reflection on graph-token understanding in future research.
% Recently, LLMs have addressed the limited generalization and narrow task scope of graph models on TAGs. GTokenLLMs are becoming a promising line of research. However, their intricate graph tokens raise doubt about their interpretability and reliability.
% In this paper, we argue that ``GTokenLLMs do not truly understand their graph tokens" by defining a formal framework for GTokenLLMs and proposing GTEval, a methodology that extends original tasks through format- and content-level transformations to assess their understanding of graph tokens. Extensive experimental results show that existing GTokenLLMs' understanding of graph tokens is still at a preliminary stage and insufficient for true comprehension, with significant potential for further improvement. We call for more retrospectives on the research of GTokenLLMs in graph token understanding.

\clearpage
\section*{Impact Statement}
This paper advances the field of machine learning by examining the limitations of graph-tokenizing large language models (GTokenLLMs). Our work aims to improve their reliability and explainability, with potential applications in social network analysis, healthcare, and e-commerce. Moreover, our work serves as a key touchstone for assessing the soundness and long-term viability of the GTokenLLM paradigm. While we do not anticipate immediate ethical risks, it remains important to monitor the deployment of such models to ensure fairness and transparency, particularly in high-stakes applications. Continued refinement of GTokenLLMs is necessary to mitigate potential biases and promote responsible use across diverse settings.
\bibliography{reference}
\bibliographystyle{icml2026}

%%%%%%%%%%%%%%%%%%%%%%%%%%%%%%%%%%%%%%%%%%%%%%%%%%%%%%%%%%%%%%%%%%%%%%%%%%%%%%%
%%%%%%%%%%%%%%%%%%%%%%%%%%%%%%%%%%%%%%%%%%%%%%%%%%%%%%%%%%%%%%%%%%%%%%%%%%%%%%%
% APPENDIX
%%%%%%%%%%%%%%%%%%%%%%%%%%%%%%%%%%%%%%%%%%%%%%%%%%%%%%%%%%%%%%%%%%%%%%%%%%%%%%%
%%%%%%%%%%%%%%%%%%%%%%%%%%%%%%%%%%%%%%%%%%%%%%%%%%%%%%%%%%%%%%%%%%%%%%%%%%%%%%%
\newpage
\appendix
\onecolumn
\section{Related Work}
In this section, we briefly discuss applications of LLMs to text-attributed graphs and existing benchmarks and evaluations of LLMs for graphs.
\paragraph{LLMs for Graphs.}
Existing applications of LLMs to TAGs can be broadly categorized into two lines of work. The first line, referred to as GTextLLMs, queries an LLM with textual representations of graphs. Early studies mainly focus on prompt design to help LLMs interpret textualized graph information, where a node is described by its textual features together with its neighborhood structure expressed in natural language~\cite{chen2024exploring,huang2023can}. Beyond directly listing neighboring information, subsequent works explore more structured textual representations, such as syntax trees~\cite{zhao2023graphtext}, random walks~\cite{tan2024walklm}, and code-like formats~\cite{wang2024instructgraph}. More recently, several studies~\cite{wang2024instructgraph,tan2024musegraph} further fine-tune open-source LLMs~\cite{llama,vicuna2023} to improve their general capabilities on graph-related tasks.
The second line of work is GTokenLLMs, which integrates real-world graphs into LLMs by encoding graph structures and textual features into token-level embeddings. As introduced in Section~\ref{sec:gtokenllm}, representative GTokenLLMs include InstructGLM~\cite{instructglm}, GraphGPT~\cite{tang2024graphgpt}, GraphTranslator~\cite{gtr}, TEA-GLM~\cite{wang2024llms}, and GOFA~\cite{konggofa}.

\paragraph{Benchmarks and Evaluations of LLMs for Graphs.}
Several survey studies~\cite{chen2024exploring,huang2023can} highlight the importance of label space design and structural information in applying LLMs to graph tasks. In addition, existing benchmarks and evaluations examine LLM capabilities across a range of graph-related problems, including graph reasoning~\cite{guo2023gpt4graph}, adversarial robustness~\cite{guo2024learning}, graph theory problems~\cite{wang2024nlgraph,fatemitalk,li2024can}, and graph pattern understanding~\cite{dai2024large}. Despite these efforts, current evaluations mainly focus on LLMs operating on textualized graphs, while a systematic evaluation of GTokenLLMs remains lacking. As a result, the understanding of graph tokens in GTokenLLMs remains underexplored.

% \section{An Introduction of Representative GTokenLLMs and How They Fit into the Framework}\label{sec:baselines}

\section{An Introduction of Representative GTokenLLMs and How They Fit into the Framework} \label{sec:gtokenllm}
In this section, we investigate six representative GTokenLLMs: LLaGA~\cite{chenllaga}, InstructGLM~\cite{instructglm}, GraphGPT~\cite{tang2024graphgpt}, GraphTranslator~\cite{gtr}, TEA-GLM~\cite{wang2024llms}, and GOFA~\cite{konggofa}. We show that these models fit into our proposed framework.

\textbf{LLaGA}~\cite{chenllaga} is a general framework that transforms graph-structured data into structure-aware token sequences and maps them into the LLM's token embedding space via a learned projector, enabling LLMs to perform graph tasks while preserving their general-purpose language capabilities.
Specifically, in stage~1, LLaGA employs an off-the-shelf text encoder, such as Sentence-BERT~\cite{reimers2019sentence} or SimTeG~\cite{duan2023simteg}, to obtain node embeddings from textual attributes, and transforms graph structures into structure-aware GE sequences using predefined templates. In stage~2, it projects GEs into GTs via a learned multilayer perceptron (MLP). Finally, these GTs are used as prefix tokens to query the LLM for node classification, link prediction, and node description tasks in stage~3. Notably, the LLM is kept frozen to preserve its comprehension capabilities and reduce training costs.

\textbf{InstructGLM}~\cite{instructglm} projects shallow node embeddings (e.g., Bag-of-Words~\cite{harris1954distributional} features on Cora) into graph tokens and integrates them with structure-aware prompt templates, enabling instruction-tuned LLMs to perform graph-related tasks.
Specifically, InstructGLM adopts shallow, task-specific graph features as GEs, which are mapped into GTs via a lightweight MLP projector to align with the LLM input space. The model further incorporates explicit, language-based descriptions of local graph structure (e.g., ``$<$node 1$>$ is connected with $<$node 2$>$ within one hop'') in the prompt template and fine-tunes the LLM with LoRA~\cite{hu2022lora} for node classification and link prediction tasks.

\textbf{GraphGPT}~\cite{tang2024graphgpt} aims to align LLMs' reasoning ability with graph-domain structural knowledge learned by a pretrained GNN, thereby improving the generalization of graph learning. Specifically, GraphGPT employs a text-grounded GNN trained with CLIP-style contrastive alignment to obtain GEs, where node features are text embeddings encoded by BERT~\cite{devlin2019bert}. These GEs are then projected into GTs via a lightweight MLP and organized into prompt templates that represent a central node together with one of its neighbors when querying the LLM.

\begin{table}[tp]
    \centering
    \caption{Example of GTextLLM prompts.}\label{tab:prompt}
    \vskip -0.1in
    \begin{tabularx}{\linewidth}{c|X}
    \toprule
        Type & Instruction \\ \midrule
        Zero-Shot & Paper: \{text feature\} Task: Please predict the most appropriate category for the paper. Your answer should be chosen from \{classes\}. \\ \hline
        1-Hop Title+Label & Input: Given the content and citation relationship of papers, determine the category of papers. Papers with citation relationships are more likely to share the same category. There are following categories:\{classes\}\{textual feature of central node and its one-hop neighbors\} Which category does Paper 1 belong to?\\ \bottomrule
    \end{tabularx}
    \vskip -0.2in
\end{table}
\textbf{GraphTranslator}~\cite{gtr} aims to bridge pretrained graph models and LLMs by translating graph embeddings into query tokens, enabling LLMs to serve as an open-ended interface for graph learning.
Specifically, GraphTranslator first adopts a pretrained GraphSAGE~\cite{hamilton2017inductive} to obtain GEs. It then employs a Translator module based on a Q-former to align these GEs with textual representations of the target node and its neighborhood, producing a set of query tokens (i.e., GTs). These query tokens are used as prefix tokens to query a frozen LLM, enabling various open-ended graph tasks while preserving the original graph model for predefined tasks.

\textbf{TEA-GLM}~\cite{wang2024llms} aligns GNN representations learned via contrastive learning with the LLM's input space, enabling cross-dataset and cross-task zero-shot graph learning.
Specifically, TEA-GLM pre-trains a GNN using contrastive learning and introduces a PCA-based feature alignment in the LLM embedding space to obtain GEs, where node features are text embeddings from BERT~\cite{devlin2019bert}. It then trains a lightweight linear projector to map GEs into a fixed number of GTs without tuning the LLM. Finally, these GTs are injected into a unified prompt template to query the LLM for node classification and link prediction in zero-shot settings.

\textbf{GOFA}~\cite{konggofa} proposes a generative graph language model by interleaving GNN layers with a frozen LLM compressor to compress graph structure and textual attributes into a fixed number of memory embeddings for downstream inference by another LLM.
Specifically, GOFA interleaves GNN layers with transformer layers of an LLM compressor~\cite{gecontext} to learn memory embeddings that capture both structural and attribute semantic information, whose final-layer outputs serve as GTs. These GTs are then directly used as query tokens for another LLM, enabling graph-conditioned generation and reasoning.

\section{GTextLLM Prompts} \label{sec:gtextllm}

We list the GTextLLM prompts used in Table~\ref{tab:prompt}, including two commonly used prompts from prior work~\cite{chen2024exploring}: ``zero-shot'', which contains the textual description of the central node, and ``1-hop'', which adds the text features of one-hop neighbors on top of the ``zero-shot'' prompt. In addition, we test four instruction types. Notably, the original instruction contains a textual description of the graph rather than graph tokens.

\section{Experiment Details of GTEval}\label{sec: exp_setting_detail}
\subsection{Dataset}
We conduct experiments on seven widely used datasets from different domains: Cora~\cite{yang2016revisiting}, Cora-70~\cite{mccallum2000automating}, Pubmed~\cite{yang2016revisiting}, and OGBN-Arxiv~\cite{hu2020open} from citation networks; OGBN-Products~\cite{hu2020open} and Book-History~\cite{yan2023comprehensive} from an e-commerce network; and WikiCS~\cite{mernyei2020wiki} from a web link network. Our data include shallow embeddings commonly used in classical methods, raw node texts, edge indices, node labels, label names, and masks for training, validation, and testing. All datasets are under the MIT License unless otherwise specified. Detailed descriptions of these datasets are provided below:

$\bullet$ \textbf{Cora} is a citation network of research papers in the computer science domain. In this dataset, each node represents a paper, and its raw text features include the title and abstract. An edge indicates a citation relationship between two papers. Each node label corresponds to the paper's category.

$\bullet$ \textbf{Cora-70} is a more comprehensive version of the Cora dataset, containing more papers and more classes than the subset used in other experiments. This dataset, formally known as the ``Cora Research Paper Classification Dataset'', provides a citation network for studying research paper classification in machine learning.

$\bullet$ \textbf{Pubmed} is a citation network of biomedical research papers. In this dataset, each node represents a paper and each edge denotes a citation relationship between two papers.

$\bullet$ \textbf{OGBN-Arxiv} is a citation network of papers and their citation relationships collected from the arXiv platform~\cite{hu2020open}. In this dataset, each node represents a paper and each edge denotes a citation relationship.

$\bullet$ \textbf{Book-History} is extracted from the Amazon-Books dataset and consists of items whose second-level label is ``History''. In this dataset, nodes represent books and edges indicate that two books are frequently co-viewed. Each node label corresponds to the book's third-level category. We use the book title and description as node text attributes. The task is to classify books into 24 and 12 categories, respectively.

$\bullet$ \textbf{OGBN-Products} is a co-purchase graph in which nodes represent product items on Amazon and edges indicate that two products are frequently co-purchased. This graph provides 47 categories.

$\bullet$ \textbf{WikiCS} is a web link network in which each node represents a Wikipedia page and each edge is a reference link between pages. For each node, the raw text includes the page title and content. Each node label corresponds to the page category.

\begin{table*}[t]
\centering
\small
\caption{Dataset statistics and the LLM used by each model.}
\vskip -0.1in
\label{tab:dataset_stats}
\begin{tabular}{cccccccc}
\toprule
Model & LLM & Dataset & Domain & \# Nodes & \# Edges & \# Classes & \makecell{Train/Val/Test (\%)} \\
\midrule

\multirow{4}{*}{LLaGA} & \multirow{4}{*}{Vicuna-7B-v1.5}
& Cora     & Citation    & 2,708   & 5,429      & 7  & 60.0/20.0/20.0 \\
&         & Pubmed   & Citation    & 19,717  & 44,338     & 3  & 60.0/20.0/20.0 \\
&         & Arxiv    & Citation    & 169,343 & 1,166,243  & 40 & 53.7/17.6/28.7 \\
&         & Products & E-commerce  & 316,513 & 19,337,722 & 39 & 8.0/1.6/90.4 \\
\midrule

\multirow{3}{*}{GraphGPT} & \multirow{3}{*}{Vicuna-7B-v1.5}
& Cora-70  & Citation    & 25,120  & 182,280    & 70 & 40.7/13.7/13.6 \\
&            & Pubmed   & Citation    & 19,717  & 44,338     & 3  & 60.0/20.0/20.0 \\
&            & Arxiv    & Citation    & 169,343 & 1,166,243  & 40 & 53.7/17.6/28.7 \\
\midrule

\multirow{3}{*}{InstructGLM} & \multirow{3}{*}{Llama-v1-7b}
& Cora     & Citation    & 2,708   & 5,429      & 7  & 60.0/20.0/20.0 \\
&           & Pubmed   & Citation    & 19,717  & 44,338     & 3  & 60.0/20.0/20.0 \\
&           & Arxiv    & Citation    & 169,343 & 1,166,243  & 40 & 53.7/17.6/28.7 \\
\midrule

\multirow{3}{*}{TEA-GLM} & \multirow{3}{*}{Vicuna-7B-v1.5}
& Cora-70  & Citation    & 25,120  & 182,280    & 70 & 40.7/13.7/13.6 \\
&          & Pubmed   & Citation    & 19,717  & 44,338     & 3  & 60.0/20.0/20.0 \\
&          & History  & E-commerce  & 41,551  & 358,574    & 12 & 60.0/20.0/20.0 \\
\midrule

\multirow{2}{*}{GOFA} & \multirow{2}{*}{Mistral-7B}
& Cora     & Citation    & 2,708   & 5,429      & 7  & 5.2/18.5/76.4 \\
&         & WikiCS   & Web link    & 11,701  & 216,123    & 10 & 5.0/15.1/50.0 \\
\midrule

GraphTranslator & ChatGLM2-6B
& Arxiv    & Citation    & 169,343 & 1,166,243  & 40 & 53.7/17.6/28.7 \\
\bottomrule
\end{tabular}
\vskip -0.15in
\end{table*}

% \begin{table*}[t]
% \caption{Statistical information of datasets.}\label{tab:stat}
% \centering
% \vskip -0.05in
% \begin{tabular}{cccccc}
% \hline
% Dataset                 & Model           & \# Nodes & \# Edges   & \# Classes & Split               \\ \hline
% \multirow{3}{*}{Cora}   & LLaGA           & 2,708   & 5,429     & 7         & 60\%/20\%/20\%         \\
%                         & InstructGLM     & 2,708   & 5,429     & 7         & 60\%/20\%/20\%         \\
%                         & GraphGPT        & 25,120  & 182,280   & 70        & 60\%/20\%/20\%         \\ \hline
% \multirow{3}{*}{Pubmed} & LLaGA           & 19,717  & 44,338    & 3         & 60\%/20\%/20\%         \\
%                         & InstructGLM     & 19,717  & 44,338    & 3         & 60\%/20\%/20\%         \\
%                         & GraphGPT        & 19,717  & 44,338    & 3         & 60\%/20\%/20\%         \\ \hline
% \multirow{4}{*}{Arxiv}  & LLaGA           & 169,343 & 1,166,243 & 40        & public split 6:2:3  \\
%                         & InstructGLM     & 169,343 & 1,166,243 & 40        & public split 6:2:3  \\
%                         & GraphGPT        & 169,343 & 1,166,243 & 40        & public split 6:2:3  \\
%                         & GraphTranslator & 169,343 & 1,166,243 & 40        & 90,941/29,799/4,000 \\ \hline
% \end{tabular}
% \vskip -0.15in
% \end{table*}
\subsection{Implementation Details}
For all GTokenLLMs, we use their original code and the default hyper-parameter settings in the authors’ implementation. The sources are listed as follows:

$\bullet$ \textbf{LLaGA}: \href{https://github.com/VITA-Group/LLaGA}{https://github.com/VITA-Group/LLaGA}

$\bullet$ \textbf{InstructGLM}: \href{https://github.com/agiresearch/InstructGLM}{https://github.com/agiresearch/InstructGLM}

$\bullet$ \textbf{GraphGPT}: \href{https://github.com/HKUDS/GraphGPT}{https://github.com/HKUDS/GraphGPT}

$\bullet$ \textbf{TEA-GLM}: \href{https://github.com/W-rudder/TEA-GLM}{https://github.com/W-rudder/TEA-GLM}

$\bullet$ \textbf{GOFA}: \href{https://github.com/JiaruiFeng/GOFA}{https://github.com/JiaruiFeng/GOFA}

$\bullet$ \textbf{GraphTranslator}: \href{https://github.com/alibaba/GraphTranslator}{https://github.com/alibaba/GraphTranslator}

\subsection{Computing Environment and Resources}
We provide the source code for our test examples in the supplementary material, and the complete source code will be made publicly available after the review. Due to file size limitations, the checkpoint of LLMs and datasets will be publicly available after the review. The implementation of the proposed GTEval utilized the PyG module.
The experiments are conducted in a computing environment with the following specifications: 

$\bullet$ \textbf{OS}: Linux ubuntu 5.15.0-102-generic. 

$\bullet$ \textbf{CPU}: Intel(R) Xeon(R) Platinum 8358 CPU @ 2.60GHz. 

$\bullet$ \textbf{GPU}: NVIDIA A800 80GB.

% \clearpage

\end{document}